\documentclass[11pt]{article}

\usepackage[preprint]{acl}

\usepackage{times}
\usepackage{latexsym}
\usepackage{multirow}
\usepackage{booktabs}
\usepackage{cleveref}
\usepackage{float}
\usepackage{enumitem}

\usepackage[T1]{fontenc}

\usepackage[utf8]{inputenc}

\usepackage{microtype}

\usepackage{inconsolata}

\usepackage{graphicx}

\title{Knowledge Distillation for Automated AI Tutor Evaluation}

\author{Tahmid Al Hannan, Diego Garcia, Alex Njoroge, Suha Al Juboori, Tarek Sakakini \\
Folsom Lake College \\
Folsom, CA, USA \\
\texttt{\{w2124644, w2079995, w2140050\}@apps.losrios.edu}, \{aljubos, sakakit\}@flc.losrios.edu}

\begin{document}
\maketitle
 \begin{abstract}
The rapid integration of Large Language Models (LLMs) into K-12 and higher education has outpaced the development of reliable methods for evaluating their pedagogical quality. As the research community starts to explore the space of automating evaluation of AI tutors, we introduce FATE (FLC AI Tutor Evaluator), a specialized 8B-parameter language model designed to evaluate AI tutors. Aligned with the four core evaluation tracks from the BEA 2025 Shared Task, our model assesses pedagogical ability across Mistake Identification, Mistake Location, Guidance, and Actionability. Because pedagogical evaluation is a specialized task with limited labeled data, we leverage knowledge distillation from a frontier LLM to generate additional supervision, yielding absolute performance gains up to 22.63 percentage points. Finally, we demonstrate FATE's utility as an automated evaluator by benchmarking instructional responses generated by popular commercial models, including ChatGPT, Claude, Gemini, and DeepSeek. On average, we have found that Gemini 2.5 Flash perfomed best (82.88\%), then ChatGPT 5.5 Instant (80.75\%), followed by DeepSeek V4 Flash (80.13\%) and Claude Sonnet 4.6 (74.00\%).
\end{abstract}

\section{Introduction}

\begin{figure}[tbp]
    \centering
    \includegraphics[width=\columnwidth]{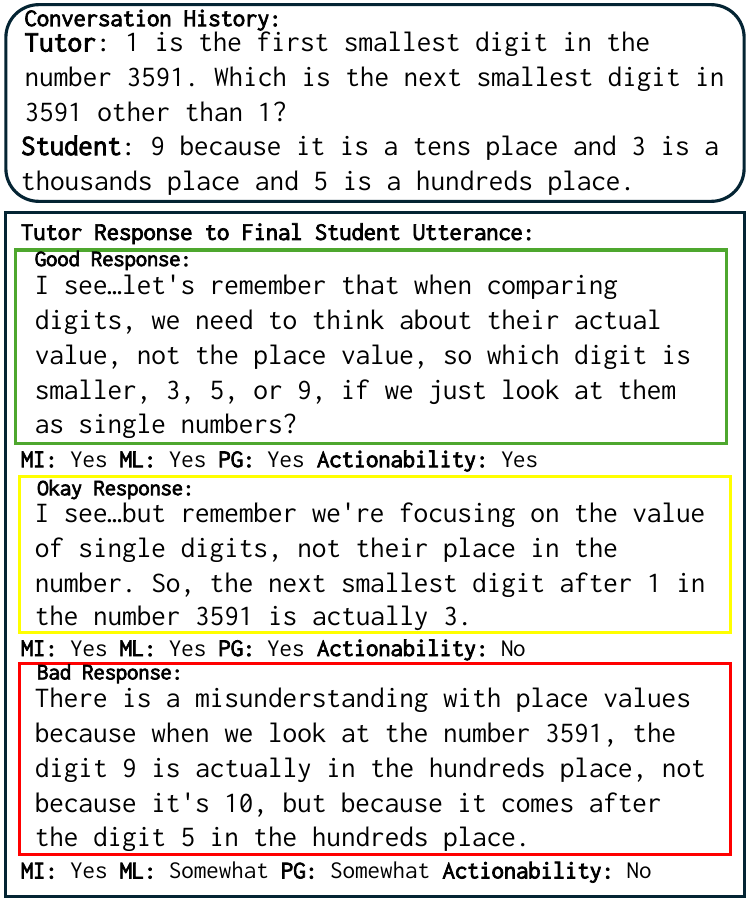}
    \caption{Example dataset point from BEA 2025. Below each AIT response is the human evaluation using four metrics: Mistake Identification (MI), Mistake Location (ML), Providing Guidance (PG), and Actionability.}
    \label{fig:datasetExample}
\end{figure}

Large language models (LLMs) are becoming increasingly prevalent in education, with AI usage in observed K--12 classrooms increasing by 84\% from 2024 to 2025 \cite{mcgeheeartificial}. As AI tutors become more widely adopted, reliable methods for evaluating their pedagogical quality are increasingly important. The BEA 2025 Shared Task introduced a benchmark for evaluating AI tutors across multiple pedagogical dimensions (Figure \ref{fig:datasetExample}) \cite{kochmar2025BEA}. However, human evaluation is costly and difficult to scale, while automatic metrics often fail to capture qualities such as guidance, actionability, and instructional effectiveness \cite{marquez2025simulation}. This motivates automated evaluators capable of assessing tutoring interactions at scale \cite{tack2023bea}.

Recent evaluation frameworks extend beyond traditional benchmarks. Multimodal environments such as TutorBench provide comprehensive evaluations but require substantial computational resources and are often mismatched with the text-based chatbot interfaces most widely used today \cite{srinivasa2025tutorbench,kajan2025stem}. Text-based tutoring has also been shown to outperform multimodal approaches in some educational settings \cite{syal2026dialogue}, motivating lightweight, specialized evaluators.

We present FATE (FLC AI Tutor Evaluator), a lightweight evaluator based on Llama 3.1 8B that assesses AI tutors across the four pedagogical dimensions of the BEA 2025 Shared Task. Trained on 1,450 student-tutor conversations using knowledge distillation from Claude Opus 4.7, FATE also enables benchmarking of commercial AI tutors including ChatGPT, Claude, Gemini, and DeepSeek.

Our contributions are:

\begin{itemize}[itemsep=0pt, parsep=0pt, topsep=0pt]
  \item Training FATE, a lightweight specialized model for AIT evaluation.
  \item Demonstrating the effectiveness of knowledge distillation for pedagogical evaluation given limited labeled data.
  \item Deploying FATE to benchmark popular commercial AI tutors.
\end{itemize}

\section{Prior Work}

Intelligent tutoring systems (ITSs) have long provided adaptive, step-level support to students, with prior work showing that ITSs can approach the effectiveness of human tutoring in some settings \cite{vanlehn2011relative}. However, LLM-based tutors differ from classical ITSs because they generate open-ended natural-language responses rather than operating over fixed solution graphs or hand-authored cognitive models. This shift makes evaluation more difficult: a response may be fluent and correct while still failing to diagnose the student's misconception, locate the error, provide useful guidance, or suggest actionable next steps.

Recent work has begun to formalize this evaluation problem. The AI Teacher Test proposed measuring whether generative models can behave pedagogically in educational dialogue \cite{tack2022ai}. The BEA 2023 Shared Task benchmarked teacher-response generation, while the BEA 2025 Shared Task reframed the problem around evaluating AI tutor responses across Mistake Identification, Mistake Location, Providing Guidance, and Actionability \cite{tack2023bea, kochmar2025findings}. Other recent benchmarks, including MRBench, TutorGym, and TutorBench, further show that tutoring quality is multi-dimensional and cannot be reduced to final-answer correctness alone \cite{maurya2025unifying, weitekamp2025tutorgym, srinivasa2025tutorbench}.

FATE is also related to broader work on LLM-based evaluation. LLM-as-judge methods have shown promise for scalable assessment of open-ended model outputs, but also exhibit biases related to position, verbosity, and model preference \cite{zheng2023judging, liu2023g}. Learned evaluators such as Prometheus and Prometheus 2 demonstrate that open models can be trained to provide rubric-conditioned judgments \cite{kim2024prometheus, kim2024prometheus2}. FATE follows this direction but specializes it for AI-tutor evaluation: rather than building a general-purpose judge, we distill supervision from a frontier model into a lightweight 8B evaluator aligned with the four BEA pedagogical dimensions.


\section{Method}
\label{sec:method}

At the center of our method is fine-tuning a pretrained lightweight LLM on the BEA 2025 Shared Task dataset incorporating heuristics from previous work \cite{fan2025bjtu}. We then leverage knowledge distillation from larger models, especially given the limited size of the original human-annotated dataset.

\subsection{Data}
\label{sec:data}

Our primary dataset is the annotated development set from the BEA 2025 Shared Task on evaluating AI tutors. It consists of 300 mathematics tutoring dialogues designed to assess the pedagogical quality of AI tutor responses rather than response generation itself \cite{kochmar2025BEA}. Each dialogue contains a multi-turn conversation ending with a student's final utterance, typically expressing a misconception, incorrect answer, or confusion. Multiple candidate tutor responses are provided for the same dialogue, generated by frontier LLM tutors (e.g., Claude, ChatGPT, Gemini, and Llama).

Each tutor response is annotated by expert evaluators across four pedagogical dimensions: \textit{Mistake Identification}, \textit{Mistake Location}, \textit{Providing Guidance}, and \textit{Actionability}. Each dimension receives one of three labels (\textit{Yes}, \textit{To Some Extent}, or \textit{No}) allowing direct comparison of different tutoring systems under identical student interactions while grounding evaluation in expert pedagogical judgments.

\begin{table*}[tb]
\centering
\begin{tabular}{llcccc}
\toprule
\textbf{Training Dataset} & \textbf{Dimension} & \textbf{Lenient F1} & \textbf{Lenient Acc.} & \textbf{Strict F1} & \textbf{Strict Acc.} \\

\midrule

\multirow{5}{*}{BEA Dataset} 
 & Mistake Identification & 84.04 & 84.98 & 62.44 & 74.17 \\
 & Mistake Location & 70.16 & 78.98 & 49.48 & 70.27 \\
 & Guidance & 68.79 & 81.68 & 48.49 & 56.46 \\
 & Actionability & 79.51 & 82.58 & 58.21 & 69.07 \\
\midrule

\multirow{5}{*}{BEA Dataset + Distillation}
 & Mistake Identification & 96.08 & 96.13 & 69.48 & 91.84 \\
 & Mistake Location & 90.58 & 90.63 & 61.52 & 80.67 \\
 & Guidance & 76.98 & 77.93 & 54.89 & 62.52 \\
 & Actionability & 83.48 & 83.72 & 55.21 & 69.89 \\
\midrule

\bottomrule
\end{tabular}
\caption{Performance metrics before and after knowledge distillation}
\label{tab:track_metrics}
\end{table*}

\subsection{Training}
\label{sec:training}

We evaluated Llama 3.2 3B, Llama 3.1 8B, and Qwen 2.5 7B, selecting Llama 3.1 8B based on overall performance. Models trained on the original BEA dataset used 3 epochs, a learning rate of $2\times10^{-4}$, and 4 gradient accumulation steps. After incorporating the knowledge distillation dataset, training used 1 epoch with the same learning rate and 16 accumulation steps. LoRA fine-tuning with 4-bit quantization enabled efficient training on A100 and T4 GPUs.

Rather than training a single evaluator, we fine-tuned four task-specific models corresponding to Mistake Identification, Mistake Location, Providing Guidance, and Actionability. Each model used prompts describing the dialogue history and evaluation rubric, following the prompting strategy of \cite{fan2025bjtu}.

To improve generalization, we adopted the dialogue-shuffling augmentation strategy proposed by \cite{fan2025bjtu}, which reorders historical dialogue turns during training. Despite preserving conversation content, this augmentation consistently improved macro F1 and accuracy.

\subsection{Knowledge Distillation}
\label{sec:knowledge_distillation}

FATE was initially trained using only the 300 student--tutor conversations from the BEA 2025 Shared Task \cite{kochmar2025BEA}. Although these conversations contained 2,480 AI tutor (AIT) and human tutor responses, the resulting model plateaued at an average macro F1 score of 74.84\%, largely due to the dataset's limited size and diversity. To improve generalization, we expanded the training set through knowledge distillation, generating synthetic student--tutor conversations with a frontier LLM. Following the findings of \cite{stanton2021does}, which show that excessively large distillation datasets can hinder optimization, we deliberately limited the amount of synthetic data while targeting our four pedagogical evaluation dimensions.

We first analyzed the label distribution of the BEA dataset and observed substantial class imbalance across all evaluation categories. For example, Mistake Identification contained 1069 \textit{Yes}, 592 \textit{No}, and 137 \textit{To Some Extent} labels. To improve balance, Claude Opus 4.7 was prompted to generate nine tutor responses for each synthetic conversation: three high-quality, three average, and three poor responses. This process produced 1,350 conversations and 12,150 labeled tutor responses. Taking Mistake Identification as an example, the distribution became substantially more balanced, increasing to 4,387 \textit{Yes}, 6,630 \textit{No}, and 1,133 \textit{To Some Extent} examples.

\begin{table}[htbp]
\centering
\resizebox{\columnwidth}{!}{
\begin{tabular}{lcccc}
\toprule
\textbf{Strategy} & \textbf{Len. F1} & \textbf{Len. Acc.} & \textbf{Strict F1} & \textbf{Strict Acc.} \\
\midrule
Baseline & 82.04 & 82.88 & 59.00 & 73.27 \\
+ Downampling & 75.53 & 75.98 & 54.95 & 66.07 \\
+ Shuffling & \textbf{84.04} & \textbf{84.98} & \textbf{62.44} & \textbf{74.17} \\
+ Shuffling + Downampling & 79.16 & 79.88 & 57.47 & 69.37 \\
\bottomrule
\end{tabular}
}
\caption{Performance of Llama 3.1 8B under different training strategies for the Mistake Identification task.}
\label{table:stratseval}
\end{table}

\section{Experimental Results}
\label{sec:results}

We first compared training strategies. Table \ref{table:stratseval} shows that dialogue shuffling consistently improved performance, whereas downsampling reduced accuracy. We report both \emph{lenient} metrics, which merge the \emph{Yes} and \emph{To Some Extent} labels, and \emph{strict} metrics, which evaluate all three labels separately.

\subsection{Impact of Knowledge Distillation}
\label{sec:distillimpact}

Using the selected Llama 3.1 8B configuration, we incorporated the synthetic knowledge distillation dataset described in Section \ref{sec:knowledge_distillation}. As shown in Table \ref{tab:track_metrics}, knowledge distillation substantially improved performance across all four pedagogical dimensions. For Mistake Identification, lenient F1 increased from 84.04\% to 96.08\%, while lenient accuracy improved from 84.98\% to 96.13\%.

\begin{figure}[ht!]
    \centering
    \includegraphics[width=\columnwidth]{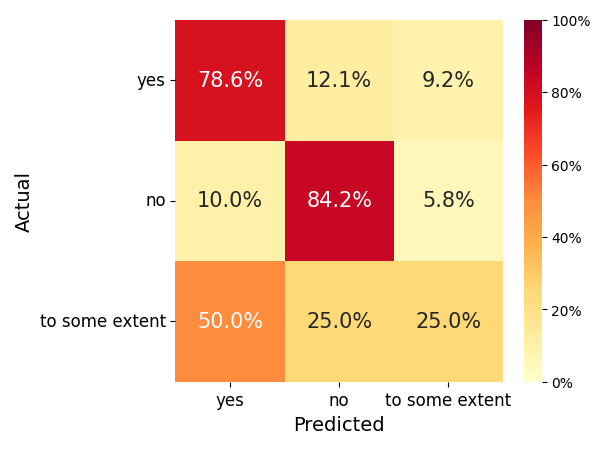}
    \caption{Llama 3.1 8B without knowledge distillation confusion matrix for Mistake Identification.}
    \label{fig:matrixoriginal}

    \vspace{1em}

    \includegraphics[width=\columnwidth]{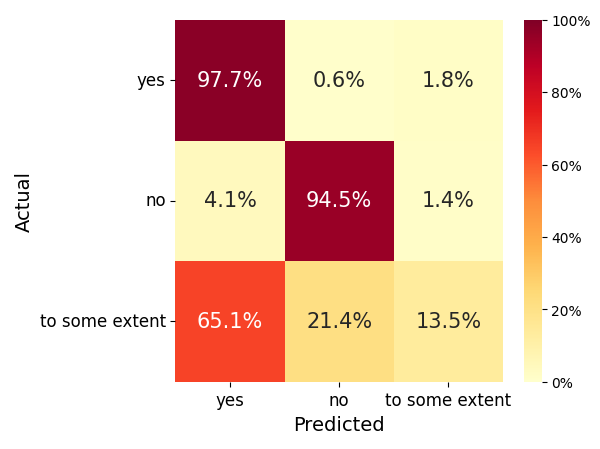}
    \caption{Llama 3.1 8B with knowledge distillation confusion matrix for Mistake Identification.}
    \label{fig:matrixdistill}
\end{figure}

To better understand the effect of knowledge distillation, we analyzed the confusion matrices for Mistake Identification before (Figure \ref{fig:matrixoriginal}) and after (Figure \ref{fig:matrixdistill}) knowledge distillation. In both settings, the model accurately classified clear \emph{Yes} and \emph{No} responses but struggled with the more ambiguous \emph{To Some Extent} category. In the baseline model, 50.0\% of \emph{To Some Extent} examples were predicted as \emph{Yes}. Although knowledge distillation further improved accuracy on clear positive (97.7\%) and negative (94.5\%) examples, it also increased this bias, with 65.1\% of ambiguous examples classified as \emph{Yes}. This behavior explains the larger gains under lenient scoring than strict scoring and suggests that compact evaluators remain challenged by nuanced, partial-credit pedagogical judgments.

\subsection{External Tutors}

To benchmark commercial AI tutors, we extracted conversation histories from the BEA 2025 dataset and generated tutor responses using Claude Sonnet 4.6, ChatGPT 5.5 Instant, Gemini 2.5 Flash, and DeepSeek V4 Flash. FATE then evaluated these responses (Table \ref{table:trackmetrics}).

Gemini achieved the highest overall average accuracy, leading in Mistake Identification, Mistake Location, and Actionability, while DeepSeek performed best on Providing Guidance. Across all models, Actionability remained the weakest category, indicating that providing concrete next-step guidance continues to be a challenge for current AI tutors.

\begin{table}[H]
\centering
\resizebox{\columnwidth}{!}{
\begin{tabular}{llcc}
\toprule
\textbf{LLM} & \textbf{Category} & \textbf{Accuracy} & \textbf{Avg. Accuracy} \\ 
\midrule

\multirow{4}{*}{Claude} 
 & Mistake ID & 91.50 & \multirow{4}{*}{74.00} \\ 
 & Mistake Location & 72.50 & \\ 
 & Guidance & 85.50 & \\ 
 & Actionability & 46.50 & \\ 
\midrule

\multirow{4}{*}{ChatGPT} 
 & Mistake ID & 98.00 & \multirow{4}{*}{80.75} \\ 
 & Mistake Location & 75.50 & \\ 
 & Guidance & 90.00 & \\ 
 & Actionability & 59.50 & \\ 
\midrule

\multirow{4}{*}{Gemini} 
 & Mistake ID & 100 & \multirow{4}{*}{82.88} \\ 
 & Mistake Location & 78.00 & \\ 
 & Guidance & 90.00 & \\ 
 & Actionability & 63.50 & \\ 
\midrule

\multirow{4}{*}{Deepseek} 
 & Mistake ID & 98.50 & \multirow{4}{*}{80.13} \\ 
 & Mistake Location & 77.50 & \\ 
 & Guidance & 90.50 & \\ 
 & Actionability & 54.00 & \\ 
\bottomrule
\end{tabular}
}
\caption{FATE's evaluation of commercially available AI Tutors.}
\label{table:trackmetrics}
\end{table}


\section{Conclusion}

We presented FATE, a lightweight LLM for automated AI tutor evaluation. Knowledge distillation substantially improved performance on this specialized task with limited labeled data, increasing lenient F1 by up to 22.63 \%. Applying FATE to commercial AI tutors further revealed a consistent weakness in actionability.

Future work will package FATE as a practical evaluation tool for educational institutions to support AI tutor adoption and assessment.

\section*{LLM Usage Acknowledgement}

We acknowledge that large language models were used to assist in the preparation of this paper, specifically for improving or polishing the language of certain sentences. All substantive ideas, analyses, and conclusions remain the responsibility of the authors.

\raggedbottom
\clearpage
\bibliography{custom}

@article{kochmar2025BEA,
  title={Findings of the bea 2025 shared task on pedagogical ability assessment of ai-powered tutors},
  author={Kochmar, Ekaterina and Maurya, Kaushal Kumar and Petukhova, Kseniia and Srivatsa, KV and Tack, Ana{\"\i}s and Vasselli, Justin},
  journal={arXiv preprint arXiv:2507.10579},
  year={2025}
}

@techreport{mcgeheeartificial,
  title={Artificial Intelligence and Student Usage in Online Learning: A Longitudinal Analysis of Usage Patterns, Achievement, and Perceptions in K-12 Virtual Education},
  author={McGehee, Nikolas},
  institution={Tech. Rep.(Michigan Virtual, 2025) accessed: 2026-02-24}
}

@inproceedings{tack2023bea,
  title={The BEA 2023 shared task on generating AI teacher responses in educational dialogues},
  author={Tack, Ana{\"\i}s and Kochmar, Ekaterina and Yuan, Zheng and Bibauw, Serge and Piech, Chris},
  booktitle={Proceedings of the 18th Workshop on Innovative Use of NLP for Building Educational Applications (BEA 2023)},
  pages={785--795},
  year={2023}
}

@inproceedings{kajan2025stem,
  title={STEM Undergraduates’ Perceptions of AI Chatbots: A Cross-Sectional Descriptive Survey},
  author={Kajan, Kamalanathan and Shi, Wenyuan and Wanatowski, Dariusz},
  booktitle={AI in Education},
  volume={1},
  number={1},
  pages={4},
  year={2025},
  organization={MDPI}
}

@article{syal2026dialogue,
  title={A Dialogue-Based Framework for Correcting Multimodal Errors in AI-Assisted STEM Education},
  author={Syal, Akshay and Prince, Lawrence Swaminathan Xavier and Gultepe, Evin and Brown, Nik Bear and Sridhar, Srinivas},
  journal={arXiv preprint arXiv:2605.04131},
  year={2026}
}

@inproceedings{fan2025bjtu,
  title={BJTU at BEA 2025 Shared Task: Task-Aware Prompt Tuning and Data Augmentation for Evaluating AI Math Tutors},
  author={Fan, Yuming and Tan, Chuangchuang and Song, Wenyu},
  booktitle={Proceedings of the 20th Workshop on Innovative Use of NLP for Building Educational Applications},
  year={2025}
}

@article{srinivasa2025tutorbench,
  title={Tutorbench: A benchmark to assess tutoring capabilities of large language models},
  author={Srinivasa, Rakshith S and Che, Zora and Zhang, Chen Bo Calvin and Mares, Diego and Hernandez, Ernesto and Park, Jayeon and Lee, Dean and Mangialardi, Guillermo and Ng, Charmaine and Cardona, Ed-Yeremai Hernandez and others},
  journal={arXiv preprint arXiv:2510.02663},
  year={2025}
}

@article{stanton2021does,
  title={Does knowledge distillation really work?},
  author={Stanton, Samuel and Izmailov, Pavel and Kirichenko, Polina and Alemi, Alexander A and Wilson, Andrew G},
  journal={Advances in neural information processing systems},
  volume={34},
  pages={6906--6919},
  year={2021}
}

@inproceedings{maurya2025unifying,
  title={Unifying AI tutor evaluation: An evaluation taxonomy for pedagogical ability assessment of LLM-powered AI tutors},
  author={Maurya, Kaushal Kumar and Srivatsa, Kv Aditya and Petukhova, Kseniia and Kochmar, Ekaterina},
  booktitle={Proceedings of the 2025 Conference of the Nations of the Americas Chapter of the Association for Computational Linguistics: Human Language Technologies (Volume 1: Long Papers)},
  pages={1234--1251},
  year={2025}
}

@article{marquez2025simulation,
  title={Simulation of teaching behaviours in intelligent tutoring systems: a review using large language models},
  author={Marquez-Carpintero, Luis and Lopez-Sellers, Alberto and Cazorla, Miguel},
  journal={Artificial Intelligence Review},
  year={2025},
  publisher={Springer}
}

@article{vanlehn2011relative,
  title={The relative effectiveness of human tutoring, intelligent tutoring systems, and other tutoring systems},
  author={VanLehn, Kurt},
  journal={Educational psychologist},
  volume={46},
  number={4},
  pages={197--221},
  year={2011},
  publisher={Taylor \& Francis}
}

@article{tack2022ai,
  title={The AI teacher test: Measuring the pedagogical ability of blender and GPT-3 in educational dialogues},
  author={Tack, Ana{\"\i}s and Piech, Chris},
  journal={arXiv preprint arXiv:2205.07540},
  year={2022}
}

@article{zheng2023judging,
  title={Judging llm-as-a-judge with mt-bench and chatbot arena},
  author={Zheng, Lianmin and Chiang, Wei-Lin and Sheng, Ying and Zhuang, Siyuan and Wu, Zhanghao and Zhuang, Yonghao and Lin, Zi and Li, Zhuohan and Li, Dacheng and Xing, Eric and others},
  journal={Advances in neural information processing systems},
  volume={36},
  pages={46595--46623},
  year={2023}
}

@inproceedings{liu2023g,
  title={G-eval: NLG evaluation using gpt-4 with better human alignment},
  author={Liu, Yang and Iter, Dan and Xu, Yichong and Wang, Shuohang and Xu, Ruochen and Zhu, Chenguang},
  booktitle={Proceedings of the 2023 conference on empirical methods in natural language processing},
  pages={2511--2522},
  year={2023}
}

@inproceedings{kim2024prometheus,
  title={Prometheus: Inducing fine-grained evaluation capability in language models},
  author={Kim, Seungone and Shin, Jay and Jang, Joel and Longpre, Shayne and Lee, Hwaran and Yun, Sangdoo and Shin, Ryan and Kim, Sungdong and Thorne, James and Seo, Minjoon and others},
  booktitle={International Conference on Learning Representations},
  volume={2024},
  pages={29927--29962},
  year={2024}
}

@inproceedings{kim2024prometheus2,
  title={Prometheus 2: An open source language model specialized in evaluating other language models},
  author={Kim, Seungone and Suk, Juyoung and Longpre, Shayne and Lin, Bill Yuchen and Shin, Jamin and Welleck, Sean and Neubig, Graham and Lee, Moontae and Lee, Kyungjae and Seo, Minjoon},
  booktitle={Proceedings of the 2024 Conference on Empirical Methods in Natural Language Processing},
  pages={4334--4353},
  year={2024}
}

@article{kochmar2025findings,
  title={Findings of the bea 2025 shared task on pedagogical ability assessment of ai-powered tutors},
  author={Kochmar, Ekaterina and Maurya, Kaushal Kumar and Petukhova, Kseniia and Srivatsa, KV and Tack, Ana{\"\i}s and Vasselli, Justin},
  journal={arXiv preprint arXiv:2507.10579},
  year={2025}
}

@inproceedings{weitekamp2025tutorgym,
  title={Tutorgym: A testbed for evaluating ai agents as tutors and students},
  author={Weitekamp, Daniel and N. Siddiqui, Momin and J. MacLellan, Christopher},
  booktitle={International Conference on Artificial Intelligence in Education},
  pages={361--376},
  year={2025},
  organization={Springer}
}




\end{document}